\documentclass[letterpaper, 10 pt, conference]{ieeeconf}  

\usepackage{cite}
\usepackage{graphicx} 
\usepackage{amsmath}
\usepackage{amssymb}
\usepackage{color}
\usepackage[font=small]{caption}
\usepackage{booktabs}
\usepackage{subcaption}
\usepackage{multirow}

\usepackage{algorithm}
\usepackage{algorithmic}
\usepackage[algo2e]{algorithm2e}

\IEEEoverridecommandlockouts                              

\title{\LARGE \bf
Dynamic Resource Management for Providing QoS \\ in Drone Delivery Systems 
}

\author{Behzad Khamidehi, Majid Raeis, and Elvino S. Sousa
\thanks{The authors are with the Department of Electrical and Computer Engineering, University of Toronto, ON M5S 1A1, Canada \hspace{2cm}
{ \tt\small  \noindent \{b.khamidehi, m.raeis, es.sousa\}@utoronto.ca}}
}

\begin{document}

\maketitle
\thispagestyle{empty}
\pagestyle{empty}

\begin{abstract}

Drones have been considered as an alternative means of package delivery to reduce the delivery cost and time. Due to the battery limitations, the drones are best suited for last-mile delivery, i.e., the delivery from the package distribution centers (PDCs) to the customers. Since a typical delivery system consists of multiple PDCs, each having random and time-varying demands, the dynamic drone-to-PDC allocation would be of great importance in meeting the demand in an efficient manner. In this paper, we study the dynamic UAV assignment problem for a drone delivery system with the goal of providing measurable Quality of Service (QoS) guarantees. We adopt a queueing theoretic approach to model the customer-service nature of the problem. Furthermore, we take a deep reinforcement learning approach to obtain a dynamic policy for the re-allocation of the UAVs. This policy guarantees a probabilistic upper-bound on the queue length of the packages waiting in each PDC, which is beneficial from both the service provider's and the customers' viewpoints. We evaluate the performance of our proposed algorithm by considering three broad arrival classes, including Bernoulli, Time-Varying Bernoulli, and Markov-Modulated Bernoulli arrivals. Our results show that the proposed method outperforms the baselines, particularly in scenarios with Time-Varying and Markov-Modulated Bernoulli arrivals, which are more representative of real-world demand patterns. Moreover, our algorithm satisfies the QoS constraints in all the studied scenarios while minimizing the average number of UAVs in use.
\end{abstract}

\section{INTRODUCTION}

The rising labor costs of the last-mile package delivery have 
encouraged many logistic companies to incorporate autonomous delivery agents (e.g., vehicles, robots, etc.) into their delivery services. In addition to reducing the delivery cost, these solutions decrease the average delivery time by extending the operational hours. Furthermore, the delivery is less affected by the availability of human resources \cite{car4pac, LastCentimeter}. Unmanned aerial vehicles (UAVs), also known as drones\footnote{Throughout the paper, we use the terms UAV and drone interchangeably.}, are among the promising options that can be used for the last-mile package delivery \cite{Informs_truck_drone, ICRA_drone_delivery_stanford, LastCentimeter, Dronet}. In this scenario, the packages are brought from the depots ( i.e., warehouses) to the package distribution centers (PDCs) by trucks or other terrestrial vehicles. Then, in the last step of the delivery chain, the UAVs deliver the packages from the PDCs to the customer locations.

Given the stochastic and time-variant nature of the package arrivals at the PDCs, it is crucial to dynamically distribute the UAVs among the PDCs such that the resource distribution matches the demand. This is in contrast to the existing methods that assume a fixed number of assigned UAVs for each PDC, which can either result in the waste of resources in low demand intervals or jeopardize the quality of service (QoS) during the peak demand periods. In order to address this issue, we study the dynamic UAV assignment problem with the goal of providing measurable QoS guarantees, while minimizing the average number of UAVs in use.
Given the practical constraints of the problem, such as the UAVs' battery limitations and the delay caused by the exchange of the UAVs between far apart PDCs, we divide the whole coverage area (e.g. city) into smaller areas called \emph{districts}, and study the dynamic UAV assignment problem in the scale of a single district (see Fig.~\ref{fig:district}). 
Furthermore, given the customer-service nature of the problem, we adopt a queueing theoretic approach to model our system, while we use a deep reinforcement learning (RL) method to avoid the limitations of the traditional queueing theoretic methods. The contributions of this paper are summarized as follows
\begin{itemize}
    \item We propose an RL-based resource management method for drone delivery application, such that the number of UAVs allocated to the distribution centers is dynamically controlled to match the demand. This is particularly important in real-world applications where the demand changes randomly through time for each PDC. This is in contrast to the existing work that consider fix UAV assignments for the distribution centers, which can lead to over-provisioning (under-provisioning) during low (high) demand intervals of a particular distribution center.
    \item Our dynamic UAV assignment scheme provides a probabilistic upper-bound on the queue length of the packages waiting to be delivered at each PDC. This is beneficial from both the service provider's and the customers' perspective. More specifically, this ensures that the distribution centers' used capacity remains below a given threshold with high probability, which is of great importance to the service provider. Moreover, since the waiting time of a package depends on the queue length upon its arrival, this upper-bound will be also beneficial to the customers as they experience bounded delays. 
    \item We use a queueing theoretic approach to model the problem. This helps us provide general insights on the resource management part of the problem, without being limited to a particular package arrival distribution, delivery location distribution, path planning method, etc. In other words, the effect of these factors will be captured by the arrival and service time distributions of our queueing models. Since our RL-based method does not make any assumptions about the arrival and service time distributions, the same approach can be applied to different problem settings.     
\end{itemize}

\section{Related Work and Background}

Different aspects of the drone delivery problem has been investigated in several recent studies \cite{Dronet, MDPI_drone_delivery, IROS_perception_aware, Pick_place_drone, IJCAI, LastCentimeter, ICRA_drone_delivery_stanford, Informs_truck_drone, TII_huang2020drone , IV_choudhury2019dynamic, TITS_joint, IoTJ_wang2019routing, Elsevier_sidekick, TITS_Synchronized, VehicleRouting}. In \cite{Dronet}, a data-driven framework has been proposed to safely drive a drone through the streets of a city. A similar problem has been investigated in \cite{MDPI_drone_delivery} and \cite{IROS_perception_aware}, where the goal is to navigate the drone based on the perception gained from its camera images. In \cite{Pick_place_drone}, the motion planning problem for a pick-up and place drone has been studied. The information exchange problem between different UAVs and its effect on the performance of a multi-UAV system has been investigated in \cite{IJCAI}. In \cite{LastCentimeter}, a personal single drone delivery system has been implemented to deliver packages to destinations within $5$km of the pickup location. In \cite{ICRA_drone_delivery_stanford, Informs_truck_drone, TITS_joint, IoTJ_wang2019routing, TII_huang2020drone , IV_choudhury2019dynamic, Elsevier_sidekick, TITS_Synchronized, VehicleRouting}, the scheduling and routing problem for a drone delivery system has been studied in different scenarios, including a complete UAV-based delivery system \cite{VehicleRouting}, a hybrid drone and truck delivery system \cite{Informs_truck_drone, TITS_joint, IoTJ_wang2019routing, Elsevier_sidekick, TITS_Synchronized}, and a drone delivery system assisted by transportation network \cite{ICRA_drone_delivery_stanford, TII_huang2020drone, IV_choudhury2019dynamic}. The focus of all the studies in \cite{ICRA_drone_delivery_stanford, Informs_truck_drone, TITS_joint, IoTJ_wang2019routing, TII_huang2020drone , IV_choudhury2019dynamic, Elsevier_sidekick, TITS_Synchronized, VehicleRouting} is on the routing of the delivery drones. To achieve this goal, two critical assumptions have been made in the aforementioned work: first, in all these studies, the demand and delivery location of all packages are known before planning. Moreover, it is assumed that each distribution center has a fixed number of UAVs, which is sufficient to satisfy its planning goal. However, in a real scenario, the demand changes randomly both temporally and spatially, and the last-mile delivery agents do not know the demand locations in advance. Moreover, the number of delivery UAVs is limited. This is particularly important during the peak demand periods or the moments that the system faces surge demand and there are fewer resources (i.e., UAVs) than required by the PDCs. To address these issues, we consider the fleet management problem for a resource-constrained drone delivery system where the number of UAVs assigned to the PDCs is dynamically adjusted to match the demand.

\subsection*{A Brief Background on Queuing Systems}
Queueing theory is the traditional method for studying the customer service systems. In these systems, customers arrive through time and need to receive a particular service by the shared servers. Because of the system's limited capacity, the customers might need to wait in a queue until a server becomes available. A single queueing system is often described using the Kendall's notation ($A/S/c$), where $A$ represents the inter-arrival time distribution, $S$ denotes the service time distribution and $c$ is the number of servers (see Fig.~\ref{fig:queue}). Although the traditional methods can provide valuable insights on the performance of the service systems, such as statistics of the waiting times or queue lengths, they often fall short on practicality because of unrealistic assumptions that are made for mathematical tractability (e.g. Poisson arrivals, exponential service times, etc.). In this paper, we follow the same approach as in \cite{queue-learning, admission-con}, where a reinforcement learning method has been used to avoid these limitations.

\subsection*{RL Background}
RL consists of the interaction between the agent and the environment in a sequence of discrete time steps $t=0,1, \ldots$ \cite{sutton2018reinforcement}. In time $t$, the environment provides the agent with state $s_t$, representing the agent's observation of the environment. The agent takes action $a_t$, receives reward $r_{t+1}$ from the environment, and goes to a new state $s_{t+1}$. To formulate this interaction, we can use a Markov Decision Process (MDP) which is denoted by $<\mathcal{S}, \mathcal{A}, \mathcal{P}, r, \gamma>$, where $\mathcal{S}$ is the finite state space, $\mathcal{A}$ is the action space, $\mathcal{P}$ is the (stochastic) transition probability function, $r$ 
is the reward function, and $\gamma \in (0,1]$ is the discount factor. The action selection mechanism is called \textit{policy}, denoted by $\pi(a|s)$, which is the probability of taking action $a$ in state $s$, i.e., $\pi (s,a) = \text{Pr} \{ a_t =a | s_t=s\}$. Let $G_t = \sum_{k=0}^{\infty} \gamma^k r_{t+k+1}$ denote the return function. We define the Q-function as $Q_{\pi}(s,a) = \mathbf{E}_{\pi} \{ G_t | s=s_t, a = a_t \}$ which shows the expected return the agent receives over the long run if it takes action $a$ in state $s$ and follows policy $\pi$ afterwards. The agent's goal is to find policy $\pi^*$ that maximizes $Q_{\pi}(s,a)$. 

\begin{figure}[t]
    \centering
    \includegraphics[trim={0.cm 0.cm 0.cm 0cm}, width=0.6\linewidth]{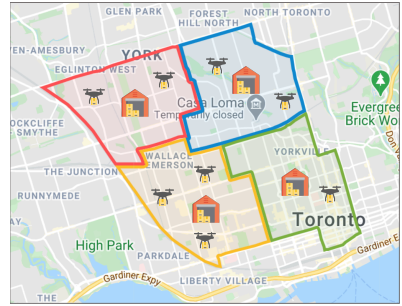}
    \caption{Considered district in downtown Toronto, including 4 regions and a total area of $36.3\text{km}^2$. The regions are created based on the city of Toronto's neighborhood profiles data set \cite{Toronto_data}.}
    \label{fig:district}
\end{figure}

\noindent \textbf{Double Deep Q-Network (DDQN)}:  Finding $Q_{\pi}(s,a)$ for every state-action pair $(s,a)$ is intractable for large state/action spaces. In these cases, we can use multi-layer neural networks to estimate the Q-function, i.e., $Q_{\pi} (s,a) \approx Q(s,a; \theta)$, where $\theta$ is the parameter of the neural network. In DDQN \cite{DDQN}, the experience tuples $(s_t, a_t, r_t,s_{t+1})$ are stored in a replay memory buffer. To update the parameters of the neural network, $\theta$, we sample a mini-batch from the experience replay buffer and minimize the loss function, defined as $\mathcal{L} (\theta) = \mathbf{E}_{\pi} \{ (y_t^{DDQN} - Q (s,a; \theta))^2\}$, 
where 
\begin{equation}
\label{DDQN}
    y_t^{DDQN}  = r_{t+1} + \gamma Q(s_{t+1},\max_a Q(s_{t+1},a; \theta) ; \theta^-).
\end{equation}
In \eqref{DDQN}, $\theta^{-}$ is the target network's parameter which is a periodic copy of $\theta$. We use the target network to stabilize the training process \cite{mnih2015human}.

\section{SYSTEM MODEL}

We consider a delivery system for a given district, which consists of $D$ regions. Each region has its own PDC that can act as a charging station as well. Let $\mathcal{D}$ and $\mathcal{N}$ denote the set of PDCs and the set of UAVs in the district, respectively. Since the UAVs are shared among all the regions, the number of UAVs assigned to each region can be dynamically changed to match the demand. The re-allocations will be carried out every $T$ time units.  
The number of UAVs allocated to the $d$-th PDC at time $t$ is denoted by $n_d (t)$. Furthermore, the district has a central UAV port for managing the UAV fleet and charging them if required. In particular, the UAVs that are not assigned to any of the regions will stay at this port until they are allocated to one of the regions. Denoting the number of unassigned UAVs in the central port by $n_0(t)$, we have $\sum_{d=0}^D n_d (t)~=~N$.

\begin{figure}
     \centering
     \begin{subfigure}{.16\textwidth}
         \centering
         \includegraphics[width=0.8\textwidth]{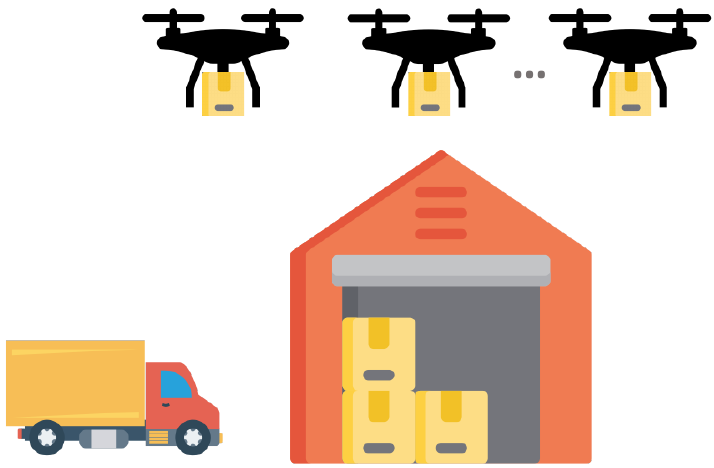}
         \caption{}
         \label{fig:depot}
     \end{subfigure}
     \begin{subfigure}{.16\textwidth}
         \centering
         \includegraphics[width=0.8\textwidth]{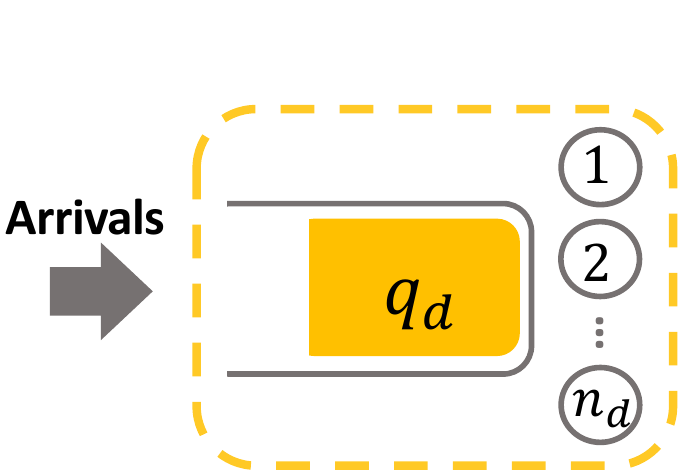}
         \caption{}
         \label{fig:queue}
     \end{subfigure}
        \caption{a) A package distribution center (PDC) with $n_d$ delivery UAVs. b) The corresponding queueing model of the PDC.}
        \label{fig:depot_queue}
\end{figure}

Given the customer-service nature of the problem, we adopt a queueing theoretic approach to model our system (see Fig.~\ref{fig:depot_queue}). In this model, the customers represent the packages which arrive at random times and need to receive a delivery service by the servers (UAVs). We model each PDC as a multi-server queueing system, where the number of servers (UAVs) need to be adjusted dynamically. Furthermore, since the packages arrive by trucks at each PDC, we use burst arrival processes to have a more realistic model. The service time captures the time required for a UAV to deliver the package to its destination and return to the assigned PDC. Moreover, the number of packages waiting to be delivered in the $d$-th queueing system (PDC) at time $t$ is represented by $q_d(t)$. These backlogged packages will be served in a First Come First Serve (FCFS) manner.

\subsection{Problem Definition}

Our goal is to provide a delivery service with minimum number of UAVs in use, while satisfying a QoS constraint for each PDC. In order to achieve this goal, the UAVs can be dynamically re-positioned between the PDCs based on the demand levels of the regions. Therefore, a key problem is to determine the number of UAVs that need to be allocated to each region at each time. As discussed earlier, while different metrics can be used as performance measures of the system, we use queue lengths of the PDCs since they can be seen as indirect measures of the system's delay. Although minimizing the queue lengths might seem appealing (especially from the customers' viewpoint), it can lead to unnecessary overuse of the resources because of the inherent trade-off between the queue length and the number of servers. On the other hand, in many delivery applications it is only required that the packages are delivered within a time window and therefore, providing an upper-bound on the queue lengths of the PDCs will be a more meaningful goal than minimizing them. Given the stochastic nature of the system (random package arrivals and delivery locations), we consider probabilistic upper-bounds on the queue lengths of the PDCs. More specifically, for each PDC we guarantee
\begin{align}
    \label{eq:qos}
    \text{P}(q_d \geq q^{ub}_d) \leq \varepsilon_d, \ \forall d \in \mathcal{D},
\end{align}
where $q^{ub}_d$ and $\varepsilon_d$ denote the probabilistic upper-bound and the violation probability, respectively. Given that the over-allocation of resources to a particular set of regions can lead to the starvation of remaining regions, we define our goal as minimizing the average number of resources in use, while satisfying the QoS constraint in \eqref{eq:qos}. 

\section{Methodology}

\begin{figure}[t]
    \centering
    \includegraphics[trim={0.cm 0.cm 0.cm 0cm}, width=0.8\linewidth]{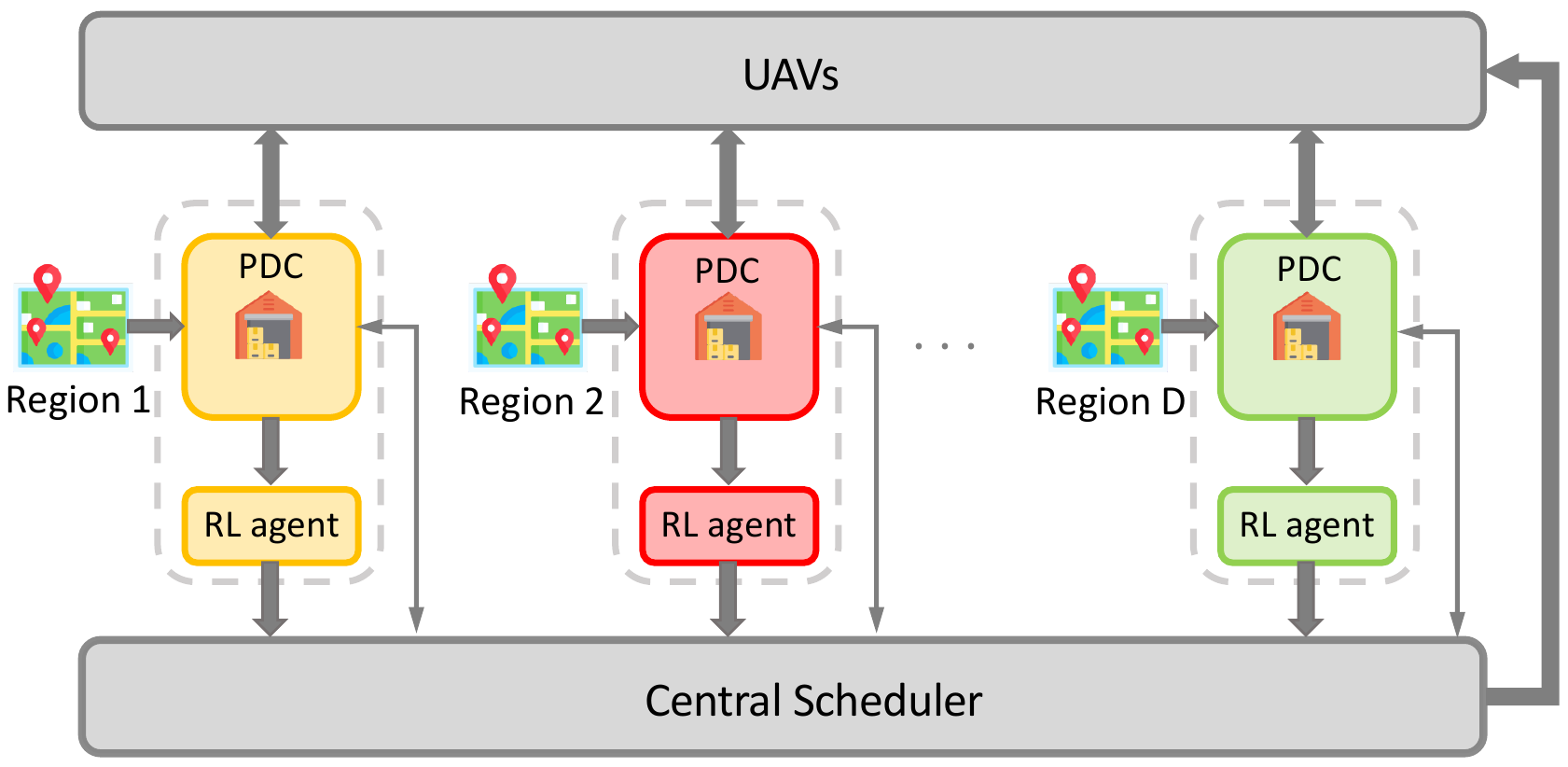}
    \caption{Block diagram of our drone delivery system.}
    \label{fig:diagram}
\end{figure}

In this section, we present our approach for achieving the goal discussed in the previous section. Fig. \ref{fig:diagram} shows the block diagram of our drone delivery system. As can be seen, our UAV management system is composed of two main components: the RL agents and the central scheduler. The RL agent of each PDC is responsible for obtaining the minimum number of UAVs required for the region to satisfy the QoS constraint in \eqref{eq:qos}, based on the real-time status of the system. However, these minimum numbers of UAVs are not  necessarily available, because of the arrival's randomness and/or the limited system's resources (UAVs). The role of the central scheduler is to handle this issue and manage coordination among the PDCs. Moreover, the scheduler determines the swapping UAVs and manages their (re)assignment to the PDCs. In what follows, we discuss how to design these two components.

\subsection{RL Agent}

To design the RL agent of each PDC, the first step is to model the problem as an MDP. In what follows, we discuss how to define the MDP for each PDC and describe its components, including the state, action, and reward function. 

\begin{itemize}
    \item \textbf{State}: We denote the state of the $d$-th agent with $s_d(t)$, which consists of the number of UAVs the PDC currently owns and its queue length, i.e., $s_d(t) = (n_d(t), q_d (t))$. 
    \item \textbf{Action}: For the $d$-th agent, we define the action $a_d(t)$ as the number of UAVs the PDC needs to take from (give to) other PDCs or the central UAV port in the next time step. 
    While we can consider a large set of actions for each PDC, without loss of generality, we restrict the action of each agent to $\{-\delta_d, 0, + \delta_d\}$, where $\delta_d \geq 1$ is an integer number. 
    \item \textbf{Reward}: Designing a proper reward function to achieve the problem's goal is its most challenging part. As mentioned earlier, our goal is to minimize the average number of UAVs in use for each region to satisfy the QoS constraint in \eqref{eq:qos}. However, there is an intrinsic trade-off between the QoS constraints and the number of UAVs in use, which must be captured by the reward function. Given that the actions are taken every $T$ time units, we define the reward function for the action taken at time $t$ as
    \begin{equation}
        \label{eq:reward}
        r_d(t) = \sum_{t': \ t' \in [t,t+T)} \tilde{r}_d(t') - n_d(t),
    \end{equation}
    where 
    \begin{equation}
        \label{eq:sub_reward}
        \tilde{r}_d(t') = \begin{cases} \alpha_1 & \text{if } q_d(t') > q_d^{ub}, \\
        \alpha_2 & \text{if } q_d(t') \leq q_d^{ub},
        \end{cases}
    \end{equation}
    for $\forall t' \in [t,t+T)$. The summation term in \eqref{eq:reward} evaluates the taken action in terms of its impact on the queue length of the PDC. The second term, however, accounts for the number of UAVs allocated to the PDC, which is fixed through the interval $[t,t+T)$. In what follows, we describe the rationale behind this reward definition and discuss how to choose the values of $\alpha_1$ and $\alpha_2$.  
    
    The goal of the RL agent is to find the optimal policy that maximizes the expected sum of rewards it receives over the long run. According to the law of large numbers (LLN), we can write
    \begin{equation}
    \label{eq:exprected_sum_reward}
        \mathbf{E}\Big[\sum_{t} r_d(t) \Big] = \lim_{K \rightarrow \infty} \frac{1}{K}\sum_{t =0}^{K} r_d(t).
    \end{equation}
    Let $k_1$ denote the number of times the queue length of the PDC is higher than the given upper bound, i.e., $q_d (t) > q_d^{ub}$. Based on the reward function in \eqref{eq:reward}, we can write 
    \begin{equation}
    \label{eq:expected_sum_rewards_2}
        \frac{1}{K}\sum_{t =0}^{K} r_d(t) = \frac{1}{K} \Big(k_1 \alpha_1 + (KT-k_1) \alpha_2 - \sum_{t=0}^{K} n_d (t) \Big).
    \end{equation}
    If we define $\alpha_1 = \varepsilon_d \lambda - \lambda$ and $\alpha_2 = \varepsilon_d \lambda$, where $\lambda$ is the reward hyper-parameter, we can simplify \eqref{eq:expected_sum_rewards_2} as
    \begin{equation}
    \frac{1}{K}\sum_{t =0}^{K} r_d(t) = -\Big( \bar{\lambda}\Big(\frac{k_1}{KT} - \varepsilon_d \Big) + \frac{1}{K} \sum_{t=0}^{K} n_d (t) \Big),
\end{equation}
where $\bar{\lambda} = \lambda T$. According to the LLN, we have $\frac{k_1}{KT} = \text{P}(q_d \geq q^{ub}_d)$ and $\frac{1}{K} \sum_{t=0}^{K} n_d (t) = \mathbf{E}[n_d(t)] $. Hence, the RL agent minimizes the following expression
\begin{equation}
\label{eq:expected_sum_rewards_3}
    \mathbf{E}[n_d(t)] + \bar{\lambda}\Big(\text{P}(q_d \geq q^{ub}_d) - \varepsilon_d \Big).
\end{equation}
We can interpret \eqref{eq:expected_sum_rewards_3} as the Lagrangian of the problem that minimizes the average number of UAVs assigned to the PDC subject to constraint \eqref{eq:qos}. The coefficient 
$\bar{\lambda}$ acts as the dual multiplier associated with the problem's constraint \cite{boyd2004convex}. This shows that our reward definition will find the policy that minimizes the average number of UAVs in use for satisfying the QoS constraint.
\end{itemize}

\subsection{Central Scheduler}

\begin{algorithm}[t]
        \small
        \SetAlgoLined
        \textbf{Initialization}: \\
        Set $\tilde{\mathcal{D}}(t)= \{ (d,a_d (t)) | d \in \mathcal{D} \ \ \text{and} \ \ a_d (t) >0\}$ and $\tilde{\mathcal{N}} = \{\}$.\\
        \textbf{Forming } $\tilde{\mathcal{N}}$:\\
        \For{ all $d \in \mathcal{D}$ with $a_d(t) <0 $}{
        Arbitrarily pick $\min (|a_d(t)|,n_d^i (t))$ idle UAVs from the PDC and add to $\tilde{\mathcal{N}}(t)$\\
        \If{$n^{i}_d (t) < |a_d(t)|$}{
        Choose the $\min (|a_d(t)|-n^{i}_d (t), n^r_d (t))$ returning UAVs that have recently finished their deliveries and add them to $\tilde{\mathcal{N}}(t)$\\
        \If{$n^{r}_d (t) < |a_d(t)|-n^{i}_d (t)$}{
        Choose the $|a_d(t)|-n^{i}_d (t) - n^{r} (t)$ delivering UAVs that have started their missions earlier and add them to $\tilde{\mathcal{N}}(t)$\\
        }
        }
        }
        \If{$\sum_{d \in \mathcal{D}} a_d (t) >0$}{
        Pick $\min (\sum_{d \in \mathcal{D}} a_d (t), n_0 (t))$ UAVs from the central port and add them to $\tilde{\mathcal{N}}(t)$}
        \textbf{Assigning UAVs to PDCs}:\\
        \While{$\tilde{\mathcal{D}}(t)\neq \varnothing$ and $\tilde{\mathcal{N}}(t) \neq \varnothing $}{
        Randomly pick PDC $d$ from the PDCs in $\tilde{\mathcal{D}}$.\\
        Form $ Z = \{dist(n,d) | n \in \tilde{\mathcal{N}}(t) \}$, where $dist(n,d)$ is the distance between UAV $n$ and PDC $d$\\
        Sort $Z$ and assign $\min (|a_d (t)|, |\tilde{\mathcal{N}}(t)|)$ closest UAVs in $Z$ to PDC $d$\\ 
        Remove assigned UAVs from $\tilde{\mathcal{N}}(t)$\\
        Remove $(d,a_d (t))$ from $\tilde{\mathcal{D}}(t)$\\
        }
        \If{ $\tilde{\mathcal{N}}(t) \neq \varnothing $}{
        Assign UAVs in $\tilde{\mathcal{N}}(t)$ to the central UAV port
        }
        \caption{Central scheduler}        
        \label{alg:CUR}
\end{algorithm}

The RL agents dynamically update the number of UAVs required by each PDC. However, they do not determine which UAVs to be swapped between the PDCs. To manage the coordination between different PDCs, a central scheduler is required. In what follows, we describe our proposed scheduler and discuss how it works.

First, let us define $\tilde{\mathcal{N}}(t)$ and $\tilde{\mathcal{D}}(t)$ as the set of UAVs to be swapped between PDCs in $[t, t+T)$ and the set of regions that require more UAVs for this time interval, respectively. Moreover, we classify the UAVs into three groups including 1) idle UAVs, which are currently at the PDCs and can start new deliveries, 2) delivering UAVs, that are currently delivering a package, and 3) returning UAVs, which have delivered their assigned packages and are returning to the PDCs. We denote the number of idle, delivering, and returning UAVs of the $d$-th PDC at time $t$ by $n^{i}_d (t)$, $n^{d}_d (t)$, and $n^{r}_d (t)$, respectively, where $n_d(t)= n^{i}_d (t) + n^{d}_d (t) + n^{r}_d (t)$.

\begin{figure}[t]
     \centering
     \begin{subfigure}{.23\textwidth}
         \centering
         \includegraphics[trim=0cm 0cm 0cm 0cm,clip, width=0.73\textwidth]{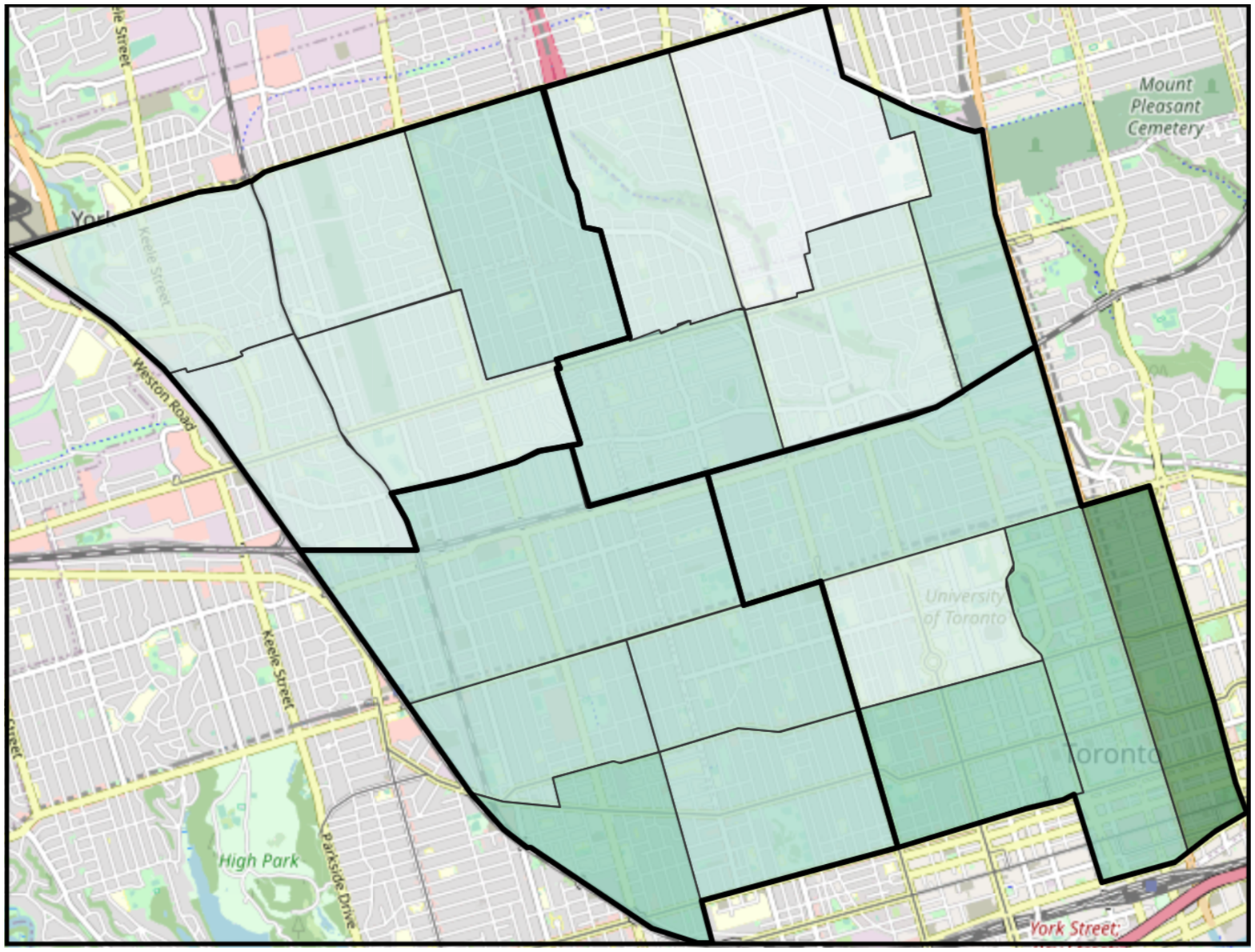}
         \caption{}
         \label{fig:}
     \end{subfigure}
     \begin{subfigure}{.22\textwidth}
         \centering
         \includegraphics[width=0.86\textwidth]{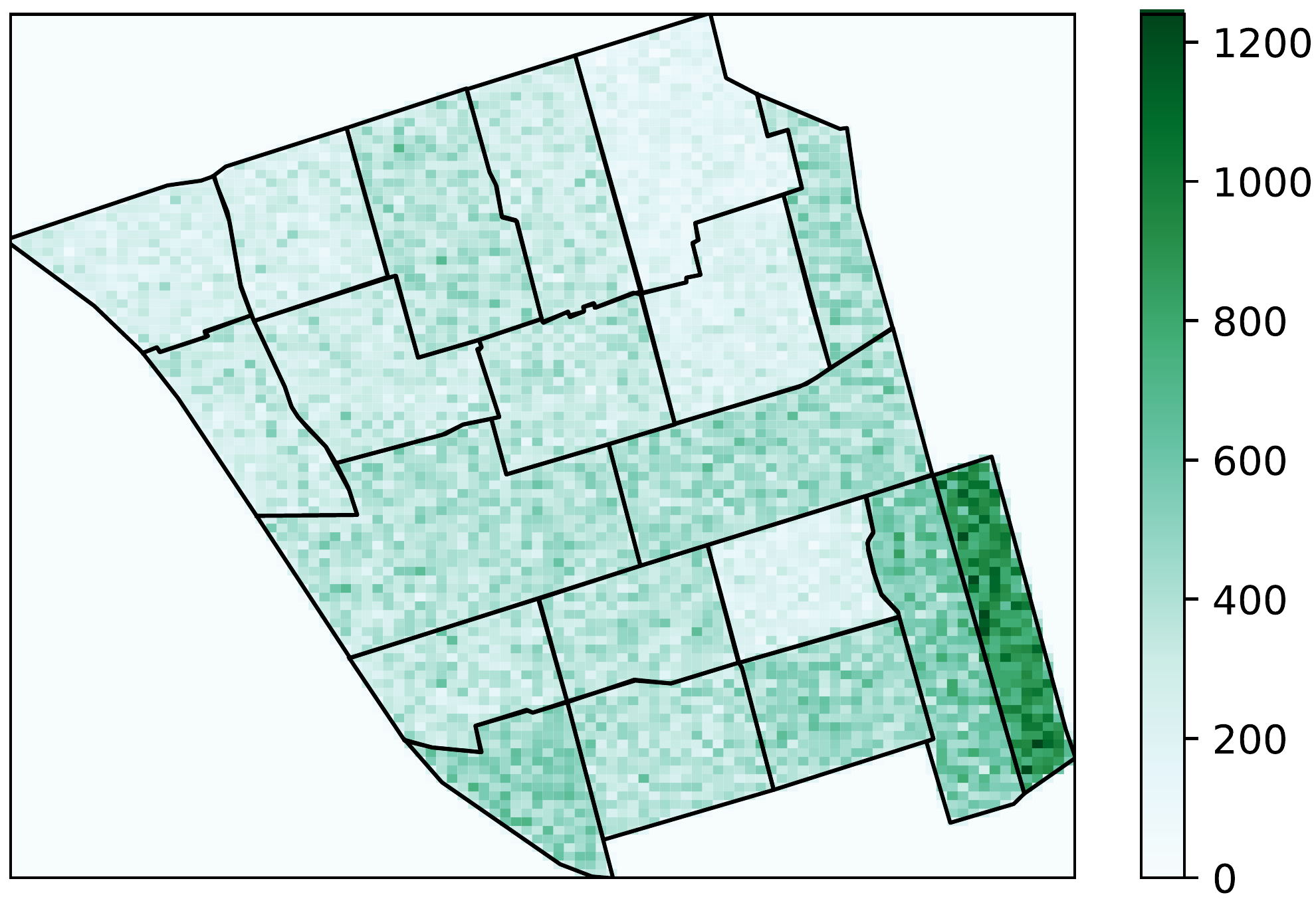}
         \caption{}
         \label{fig:}
     \end{subfigure}
        \caption{ (a) The population density, based on the Toronto's neighbourhood profile data set. (b) Histogram of the considered demand locations for the training of our RL agents.}
        \label{fig:density}
\end{figure}

\noindent Choosing the swapping UAVs is equivalent to forming set $\tilde{\mathcal{N}}(t)$. The PDCs with $a_d (t) >0$ do not give any UAVs to other PDCs. Hence, to form $\tilde{\mathcal{N}}(t)$, it is sufficient to consider the central port and the PDCs with $a_d(t) <0$. For the PDCs with $a_t^d <0$, we start with the idle UAVs. If $n^{i}_d (t) \geq |a_d(t)|$, we arbitrarily pick $|a_d(t)|$ idle UAVs from the PDC and add them to $\tilde{\mathcal{N}}(t)$. Otherwise, we select all $n^{i}_d (t)$ idle UAVs and add them to $\tilde{\mathcal{N}}(t)$. For the remaining $|a_d(t)|-n^{i}_d (t)$ UAVs, the scheduler considers the returning UAVs first. If $n^{r}_d (t) \geq |a_d(t)|-n^{i}_d (t)$, the scheduler chooses the $|a_d(t)|-n^{i}_d (t)$ UAVs that have recently finished their deliveries, and adds them to $\tilde{\mathcal{N}}(t)$. 
If $n^{r}_d (t) < |a_d(t)|-n^{i}_d (t)$, the scheduler selects all $n^{r}_d (t)$ returning UAVs and adds them to $\tilde{\mathcal{N}}(t)$. For the rest of swapping UAVs, the scheduler picks $|a_d(t)|-n^{i}_d (t) - n^{r} (t)$ UAVs from those that are currently delivering the packages. These UAVs must deliver their packages first and then go to the new region. In this step, we select the UAVs that have started their missions earlier. 
The last step to form set $\tilde{\mathcal{N}}(t)$ is to check $ \delta (t) = \sum_{d \in \mathcal{D}} a_d (t)$. If $\delta (t) >0$ and there are enough UAVs in the central UAV port, the scheduler adds $\delta (t)$ UAVs to $\tilde{\mathcal{N}}(t)$. Otherwise, the scheduler adds all $n_0 (t)$ UAVs to $\tilde{\mathcal{N}}(t)$ and neglects the remaining required ones as there is no more UAV in the district.

\noindent After forming $\tilde{\mathcal{N}}(t)$, the scheduler must assign these UAVs to the PDCs in $\tilde{\mathcal{D}}(t)$. To achieve this goal, the scheduler randomly chooses a PDC from $\tilde{\mathcal{D}}(t)$. Then, it evaluates the distance between the current position of the UAVs and the considered PDC (for delivering UAVs, this is equal to the distance between the UAV and its destination plus the distance between the destination and the new PDC). The scheduler chooses $|a_d(t)|$ closest UAVs and assigns them to the $d$-th PDC. This procedure is repeated until all UAVs are assigned to the new PDCs (or the central port). The random PDC selection for the UAV assignment is for providing fairness to all PDCs. The pseudo-code of this procedure is given in Algorithm I.

\section{Evaluation and Results}
In this section, we evaluate the performance of our proposed algorithm for a drone delivery system.

\subsection{Experimental Setup and Data set}

We implement our drone delivery environment in Python. For our environment, we consider a district in downtown Toronto, with a total area of $36.3\text{km}^2$, as depicted in Fig.~\ref{fig:district}. The district includes $4$ regions. To create the regions, we use the city of Toronto's neighborhood profiles data set \cite{Toronto_data}, which contains the geographical characteristics of the city of Toronto. For each region, we consider a center (PDC) located roughly in the middle of the region. Moreover, we consider a UAV port in the middle of the district. The UAVs that are not used by any of the PDCs will stay at this port until they are allocated to a new region. 

\noindent The average speed of the drone is set to $18$kph, based on the DJI Mavic pro specifications\footnote{https://www.dji.com/ca/mavic/info}. As the drone's flight time is limited, its battery will be replaced with a fully charged one after each delivery mission once it arrives at its assigned PDC. The drone is unavailable during the battery replacement period. However, this does not have much effect on the performance of the system, as the battery replacement period is negligible compared to the delivery times.

\begin{table}[t]
    \centering
        \caption{DDQN hyper-parameters}
        \begin{tabular}{ l  l }
        \toprule
            Parameter & Value \\
            \hline 
            Adam optimizer learning rate &  0.001\\
            replay memory buffer size & 1,000,000 \\
            mini-batch size & 25 \\
            target network update frequency & 5 episodes\\
            discount factor for target & 0.99 \\
            Number of hidden layers & 2\\
            Size of hidden layers & (32, 32) \\
            maximum training steps per episode & 1000\\
            Decaying for the $\epsilon$-greedy algorithm & $0.5$ to $0.05$\\
            \bottomrule
        \end{tabular}
    \label{table:hyper_parameters}
\end{table}

Package arrivals are generated based on the demo-graphical data from the city of Toronto~\cite{Toronto_data}. We have used this data to obtain the demand rates for each region, as shown in Fig.~\ref{fig:density}. Since the packages are brought to each PDC by trucks, we use batch arrivals to have a realistic model of the system. Moreover, three different package arrival distributions have been considered to represent different demand patterns through time: Bernoulli, Time-Varying Bernoulli (TVB), and Markov-Modulated Bernoulli (MMB) processes. In the Bernoulli arrival process, we can have either one or zero batch arrivals at a given time slot with probabilities $p$ and $1-p$, respectively. This is the discrete-time version of the Poisson process.  While Bernoulli (Poisson) process has been widely used in the literature to represent customer arrivals, it does not capture the time-varying nature of the arrivals. Therefore, we use the Time-varying Bernoulli process, in which the probability of arrivals ($p$) changes as a function of time. Specifically, we consider a periodic pattern of high and low demand intervals, during which the probability of a batch arrival in a given time slot is $p_{high} = 0.9$ and $p_{low} = 0.1$, respectively. The third arrival pattern (Markov-modulated Bernoulli process) includes similar high and low demand intervals, with the difference that a two-state Markov model governs the transition between these two regimes. The transition probabilities from the high-demand state to the low-demand state and vice versa are respectively equal to $\text{p}_{\text{high} \to \text{low}} = 0.15$ and $\text{p}_{\text{low} \to \text{high}} = 0.15$.

\noindent In order to generate the package destinations in each region, first a sub-region is chosen based on the Categorical distribution, where the probability of picking each sub-region is proportional to its population. Then the final destination of the package is sampled from a uniform distribution within the chosen sub-region. Fig.~\ref{fig:density}b shows the distribution of the generated destinations, which match the population density in Fig.~\ref{fig:density}a.


\begin{figure}
     \centering
     \begin{subfigure}{.23\textwidth}
         \centering
         \includegraphics[width=1\textwidth]{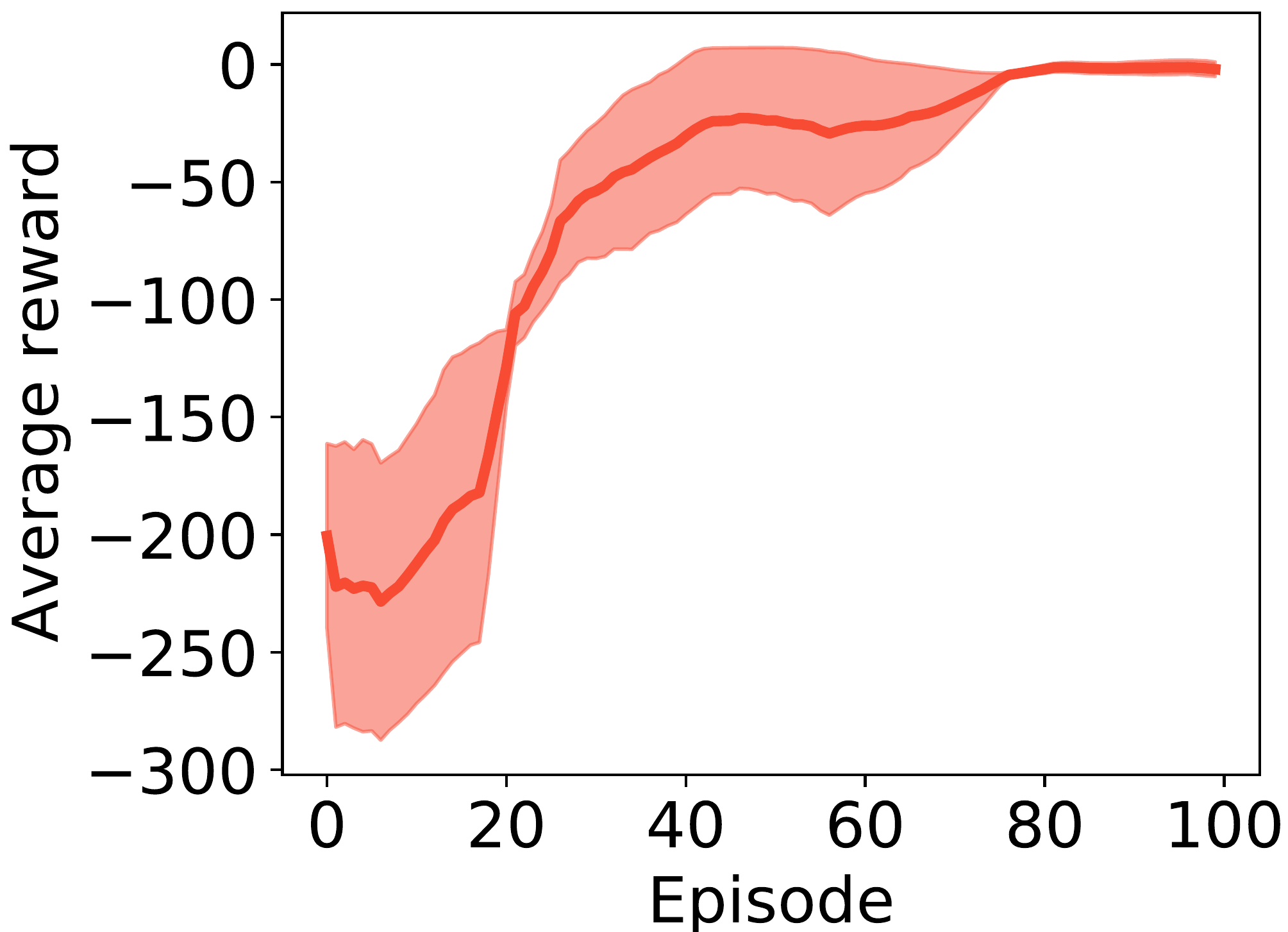}
         \caption{}
         \label{fig:Reward_training}
     \end{subfigure}
     \begin{subfigure}{.22\textwidth}
         \centering
         \includegraphics[width=1\textwidth]{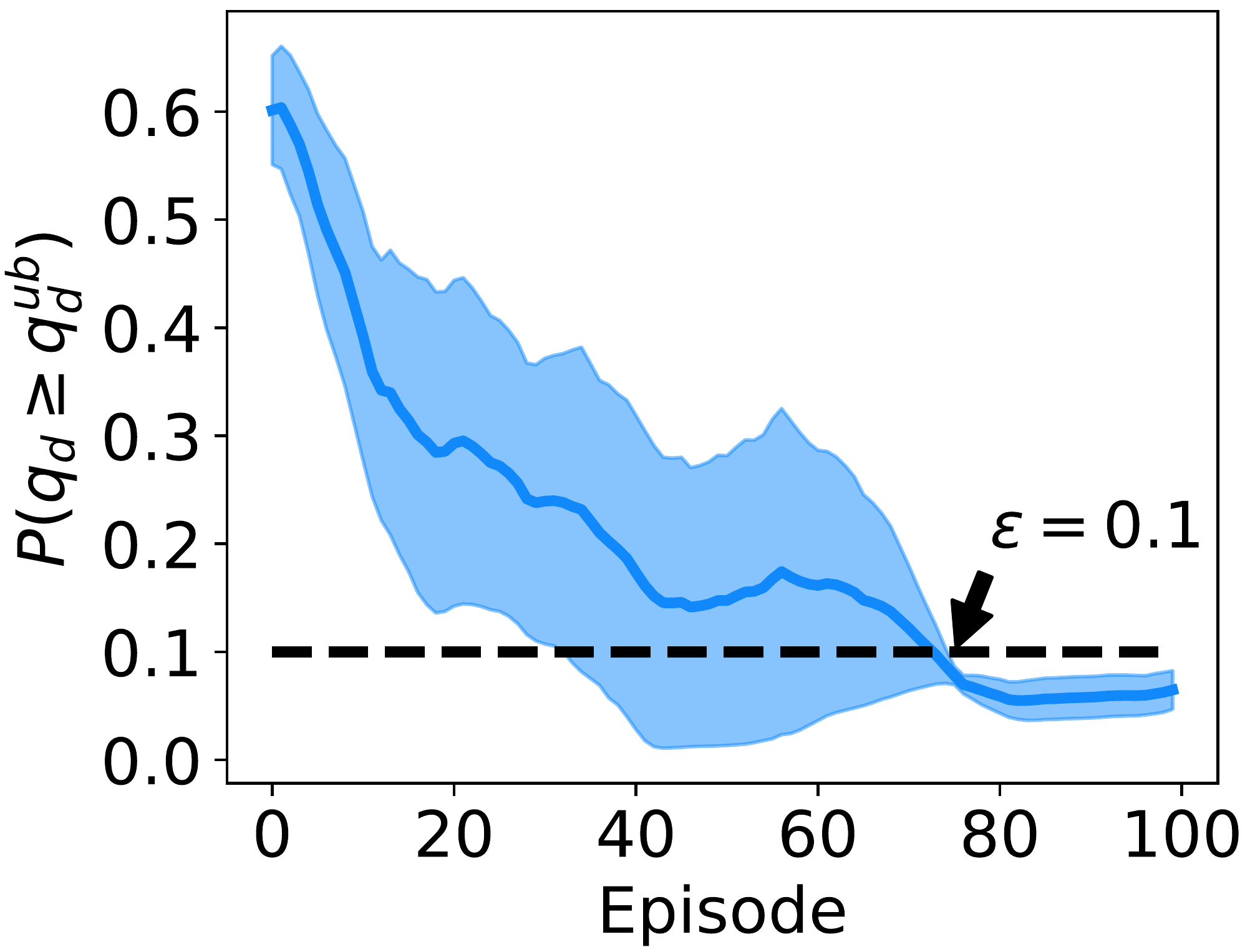}
         \caption{}
         \label{fig:Pr_training}
     \end{subfigure}
        \caption{The training curves of the algorithm, including (a) the average reward and (b) the probability of $\text{P}(q_d \geq q^{ub}_d)$. }
        \label{fig:Re_Pr}
\end{figure}

\begin{table}[t]
    \centering
        \caption{Experiment parameters.}
        \begin{tabular}{lp{3.5cm}l}
        \toprule
        Parameter &  & Value\\
        \hline
        \multicolumn{2}{l}{$\lambda$ (reward hyper-parameter)} & $4$\\
        $\varepsilon_d$, $\forall d$ & & $0.1$ \\
        $\delta_d$ & & 5\\
        \multicolumn{2}{l}{$T$ (time between two actions)} & $60$ mins\\
        \hline
        $q_d^{ub}$ & Bernoulli (Poisson) & $[85,80,120,150]$ \\
         & Time-varying  Bernoulli & $[120, 110, 165, 200]$ \\
         & Markov-modulated Bernoulli & $[120, 110, 165, 200]$ \\
         \hline
        \multicolumn{2}{l}{Number of packages per truck arrival at the } &  \\
        \multicolumn{2}{l}{$d$-th PDC $\sim \text{uniform}[x_d-15, x_d+15]$,} & \\
        \multicolumn{2}{l}{where $x_d$ is } & $[55, 50, 75, 90]$\\
        \hline
         \multicolumn{2}{l}{High and low demand arrivals period} & $300$mins \\ 
         \multicolumn{2}{l}{Trucks' inter-arrival time} & $30$mins \\
         \multicolumn{2}{l}{Battery replacement time for each UAV} & $1$min \\ 
            \bottomrule
        \end{tabular}
    \label{table:other_parameters}
\end{table}

\subsection{Implementation Parameters}

We use DDQN for the training of our RL-agents. For the neural network, we consider a fully connected neural network with two hidden layers. The hidden layers are activated by rectified linear unit (ReLU) activation functions. The last layer of the network is of size $3$ and it has no activation as it estimates the Q-function. During the training, each agent implements the $\epsilon$-greedy policy to take its action. The value of $\epsilon$ is decayed from $0.5$ to $0.05$. The actions of the PDCs are $\{-5,0,+5\}$. Unless otherwise stated, the value of $T$, which is the time interval between two consecutive actions is $60$ minutes. It is worth mentioning that the neural networks are insensitive to close decimal numbers, Hence, if we directly feed the states which include two decimal numbers, the neural network does not learn efficiently. To resolve this issue, we convert the state vector from the decimal form to the equivalent binary representation. We represent the queue length and the number of UAVs with $15$ and $10$ bits, respectively.  
The maximum training steps per episode is set to $1000$. To increase the training efficiency, we terminate each episode if the queue length of a PDC exceeds $2000$. This is particularly important in the beginning of the training as the RL agents take random actions to explore different states. These random actions can saturate the system and the RL agents cannot fix this situations.  The details of the hyper-parameter values used for the training of the DDQN algorithm are given in Table \ref{table:hyper_parameters}.

\subsection{Baselines}
We compare our proposed scheme with the following baselines:
\subsubsection{Static} In this scheme, the number of UAVs assigned to each PDC is fixed and proportional to the population density of that region. These densities are derived using the city of Toronto's demo-graphical data \cite{Toronto_data}.

\subsubsection{Threshold-based} In this method,
the number of UAVs assigned to each PDC is adjusted based on the queue length thresholds of the PDCs. If $q_d (t) < 0.5 q_d^{ub}$, the number of UAVs assigned to the PDC decreases by $\delta_d$. If $q_d(t) \geq 1.5 q_d^{ub}$, the number increases by $\delta_d$. Otherwise, the number remains unchanged.

\subsubsection{ Queue Length (QL)-based} In this scheme, the number of UAVs assigned to the PDCs is proportional to their queue lengths. We update this proportional assignment every $5T$ time units. We avoid frequent updates as it can increase the wasted time spent on the UAV swaps.



\begin{figure}[t]
    \centering
    \includegraphics[trim={0.cm 0.cm 0.cm 0cm}, width=0.57\linewidth]{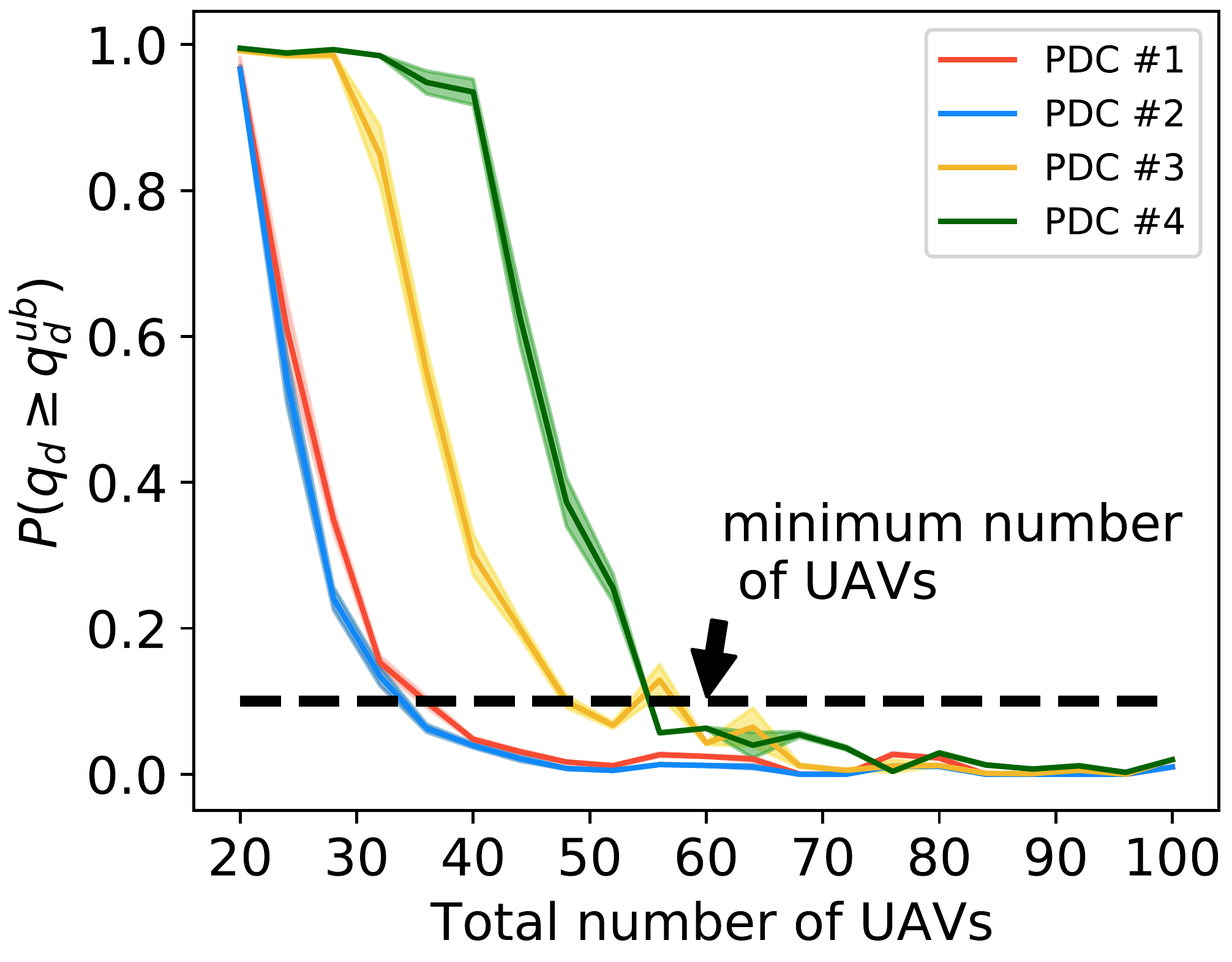}
    \caption{Impact of the total number of UAVs on $\text{P}(q_d~\geq~q^{ub}_d)$. }
    \label{fig:Pr_Total_UAVs}
\end{figure}

\begin{figure*}
     \centering
     \begin{subfigure}{.32\textwidth}
         \centering
         \includegraphics[width=0.9\textwidth]{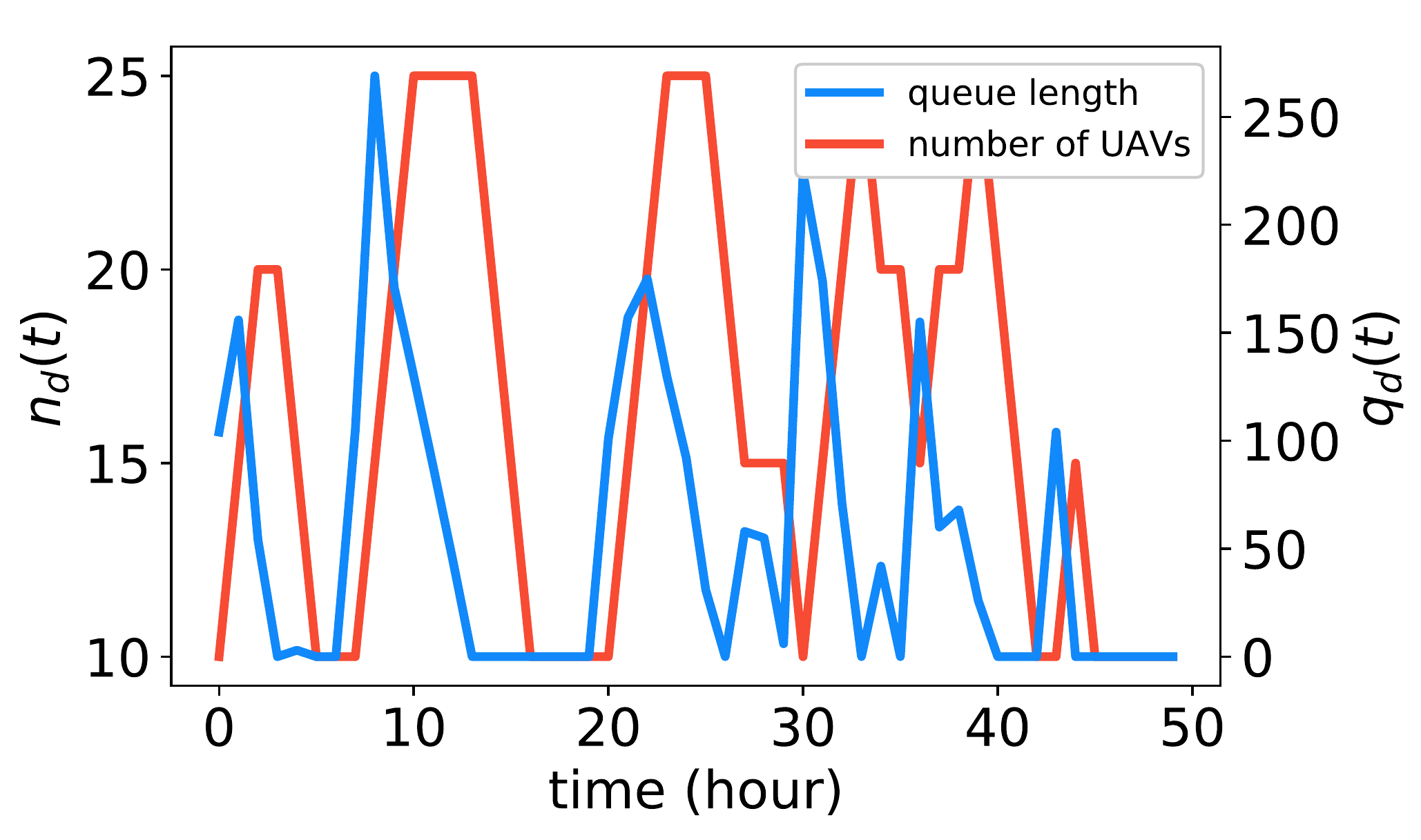}
         \caption{}
         \label{fig:n_q_1}
     \end{subfigure}
     \begin{subfigure}{.32\textwidth}
         \centering
         \includegraphics[width=0.9\textwidth]{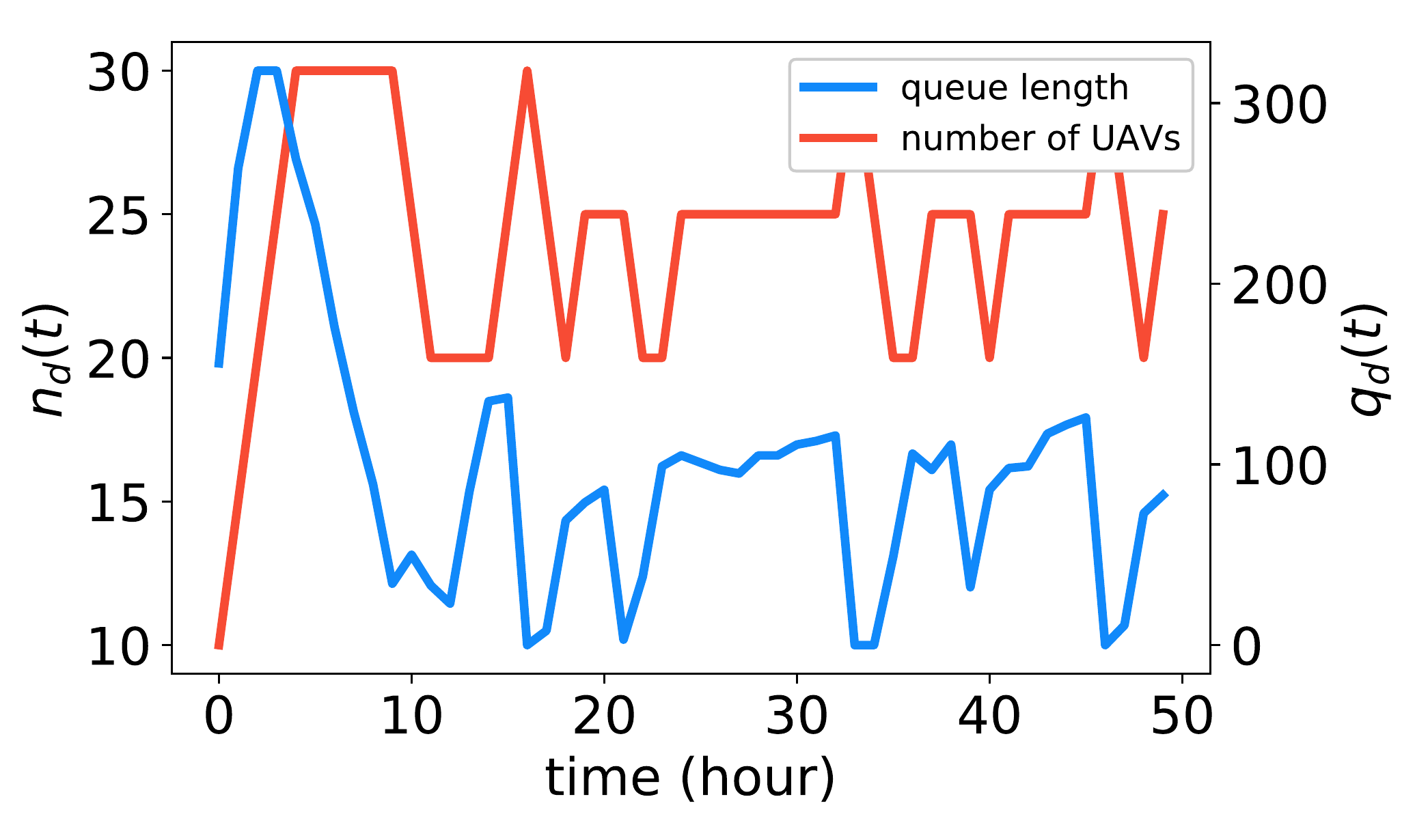}
         \caption{}
         \label{fig:n_q_2}
     \end{subfigure}
     \begin{subfigure}{.32\textwidth}
         \centering
         \includegraphics[width=0.9\textwidth]{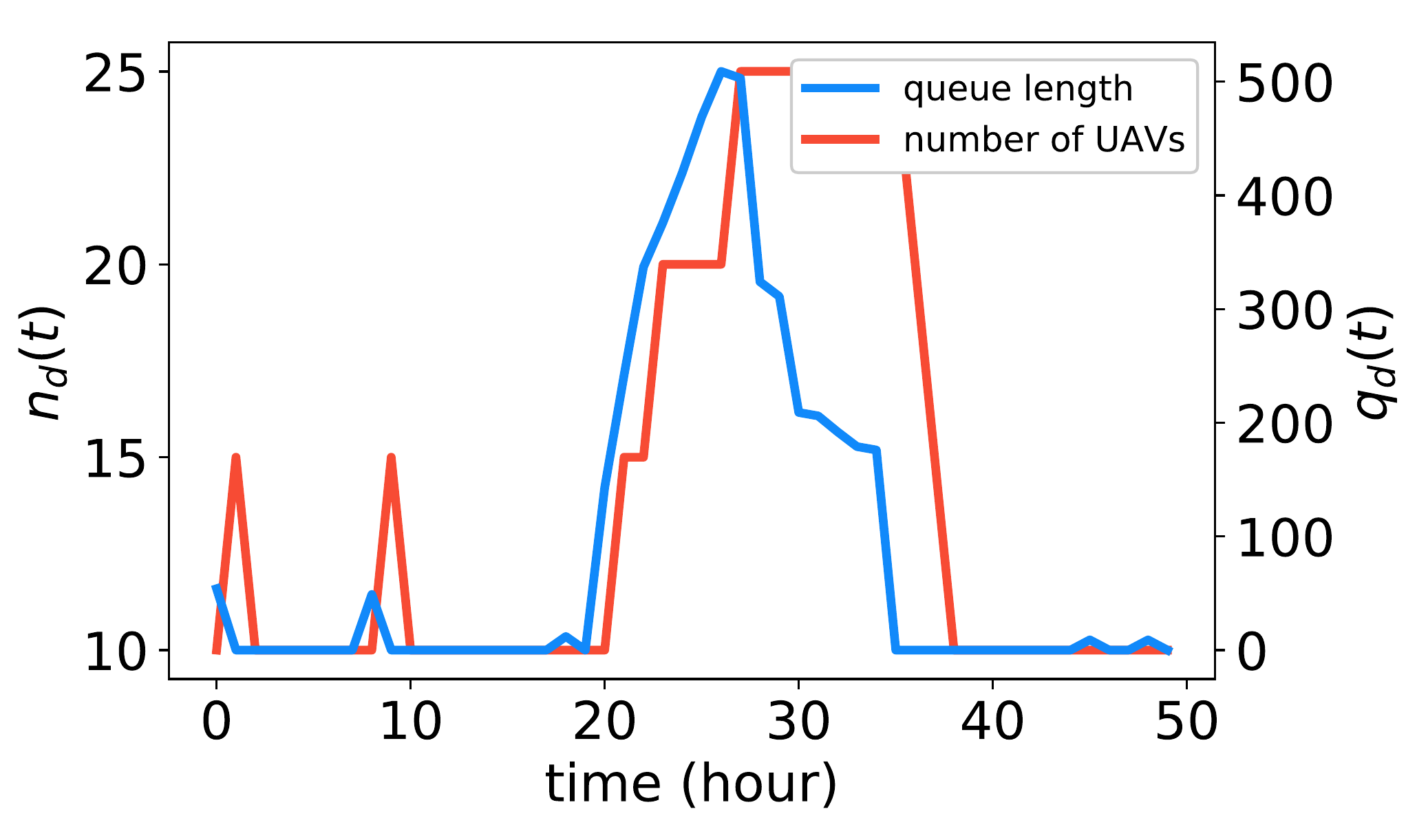}
         \caption{}
         \label{fig:n_q_3}
     \end{subfigure}
        \caption{Adjustments of the number of UAVs by our algorithm for a given sample path of the arrivals: (a) Bernoulli, (b) TVB (c) MMB. }
        \label{fig:n_q}
\end{figure*}

\subsection{Results and Discussion}

Let us first discuss the training process of our UAV management system. As mentioned earlier, our goal is to learn policies that minimize the average number of UAVs in use, while  satisfying a set of QoS guarantees for the PDCs, i.e., $P(q_d>q_d^{th})\leq \varepsilon_d$. In this experiment, we aim to bound the violation probabilities of all the PDCs by $10\%$, i.e., $\varepsilon_d=0.1, \forall d \in \mathcal{D}$, where the queue-length thresholds and other parameters of the experiment are summarized in Table \ref{table:other_parameters}. We start with sufficiently large number of UAVs to ensure the feasibility of the problem, as well as having large enough state space for exploration. However, the RL-agents learn to minimize the number of required UAVs to satisfy the constraints, as they interact with the environment. Fig.~\ref{fig:Re_Pr} shows the training curves of the average reward and the violation probability, as the RL-agents learn to minimize the number of required UAVs for providing the QoS constraint. The agents were trained with 5 different seeds. The solid curves and the pale regions show  the  average  and  the  standard  error  bands,  respectively. As can be observed, the agent learns to adjust the number of UAVs such that the violation probability decreases from its initial value (around $0.6$) and converges to a value just below $0.1$, as desired.

Now, 
we can obtain the minimum number of UAVs required to guarantee the QoS constraints by evaluating the performance of the agents for a range of different number of UAVs. As shown in Fig.~\ref{fig:Pr_Total_UAVs}, the violation probabilities of the PDCs start to increase as we shrink the number of UAVs. Furthermore, we can observe that the minimum number of UAVs required to satisfy $P(q_d>q_d^{th})\leq 0.1$ for all the PDCs is around $60$.

\subsection*{Comparisons}  In the rest of this section, we compare the performance of our proposed algorithm with the baselines introduced earlier. We conduct these comparisons under three different scenarios, where the effect of each demand pattern on the performance of the algorithms is studied. To simplify the notation, we define $p^{\max}$ as $\displaystyle \max_{d \in \mathcal{D}} P(q_d (t) > q_d^{ub})$, which represents the violation probability of the PDC with the worst performance. The average queue length, waiting time and number of used UAVs among all the PDCs are denoted by $\bar{q}$, $\bar{w}$ and $\bar{n}$, respectively. The standard deviation of the queue lengths and the waiting times among all the PDCs are represented by $\sigma_q$ and $\sigma_w$, respectively.

\begin{table}[ht]
\centering
\caption{Comparison between different algorithms. }
\begin{tabular}{l|l|p{1.2cm}p{1.3cm}p{1.3cm}}
    \toprule
    & & \multicolumn{3}{c}{ Demand Pattern}\\
    \cline{3-5}
    Algorithm & Metric & Bernoulli (Poisson) & Time-varying  Bernoulli (Poisson) & Markov-modulated Bernoulli (Poisson) \\
    \hline
    \multirow{3}{*}{Static} & $p^{\max}$  & $\mathbf{0.05}$ & $0.99$ & $0.49$\\
    & $\bar{q}$ & $\mathbf{17.7}$ & $\infty$ & $167.0$ \\
    & $\bar{w}$ & $\mathbf{34.6}$ & $\infty$ & $182.8$\\
    & $\sigma_w$ & $\mathbf{36.5}$ & $\infty$ & $479.2$ \\
    & $\sigma_q$ & $\mathbf{50.6}$ & $\infty$ & $767.5$ \\
    & $\bar{n}$ & $60$ & $60$ & $60$\\
    \hline
    \multirow{3}{*}{Threshold-based} & $p^{\max}$ & $0.51$ & $0.56$& $0.51$ \\
    & $\bar{q}$ & $119.6$ & $142.5$ & $173.9$ \\
    & $\bar{w}$ & $186.5$ & $424.8$ & $223.1$ \\
    & $\sigma_w$ & $129.9$ & $739.4$ & $304.2$ \\
    & $\sigma_q$ & $142.2$& $147.4$ & $292.5$ \\
    & $\bar{n}$ & $36$ & $40$ & $40$\\
    \hline
    \multirow{3}{*}{QL-based} & $p^{\max}$ & $0.51$ & $0.56$ & $0.41$\\
    & $\bar{q}$ & $142.2$ & $198$ & $176.5$\\
    & $\bar{w}$ & $220.1$ & $256.4$ & $252.5$\\
    & $\sigma_w$ & $207.5$ & $208.5$ & $200.6$\\
    & $\sigma_q$ & $208.8$ & $382.1$ & $317.0$\\
    & $\bar{n}$ & $60$ & $60$ & $60$ \\
    \hline
    \multirow{3}{*}{Ours} & $p^{\max}$ & $\mathbf{0.10}$ & $\mathbf{0.10}$ & $\mathbf{0.10}$ \\
    & $\bar{q}$ & $29.3$ & $\mathbf{32.8}$ & $\mathbf{38.4}$\\
    & $\bar{w}$ & $49.9$ & $\mathbf{43.4}$ & $\mathbf{60.8}$\\
    & $\sigma_w$ & $45.8$ & $\mathbf{48.8}$ & $\mathbf{68.7}$ \\
    & $\sigma_q$ & $66.9$ & $\mathbf{67}$ & $\mathbf{107.5}$ \\
    & $\bar{n}$ & $\mathbf{50.4}$ &  $\mathbf{57.3}$ & $\mathbf{53.8}$ \\
    \bottomrule
\end{tabular}
\label{tab:comparison}
\end{table}

\subsubsection{Bernoulli Arrivals} As mentioned earlier, this is the discrete time version of the well-known Poisson process, where an arrival can occur in a given time step with probability $p$. Considering Bernoulli arrivals for all the regions, the performance of our algorithm and the baselines are summarized in Table \ref{tab:comparison}. As can be observed, our algorithm and the static assignment both satisfy the QoS constraint, i.e., $p^{\max}\leq 0.1$. While the static algorithm results in smaller queue lengths and waiting times, ours minimizes the average number of UAVs in use. In other words, our algorithm exactly does what it was designed for. Therefore, it uses just enough resources (UAVs) to satisfy the QoS guarantees, instead of consuming all the available resources as in the static assignment. Fig.~\ref{fig:n_q_1} shows the dynamic adjustment of the number of UAVs by our algorithm, along with the variation of the queue length. Furthermore, it seems that the threshold-based algorithm under-uses the available resources and therefore, cannot satisfy the QoS constraint. QL-based algorithm cannot satisfy the constraints either, although it uses all the available resources.

\begin{figure}[t]
    \centering
    \includegraphics[trim={0.cm 0.cm 0.cm 0cm}, width=0.6\linewidth]{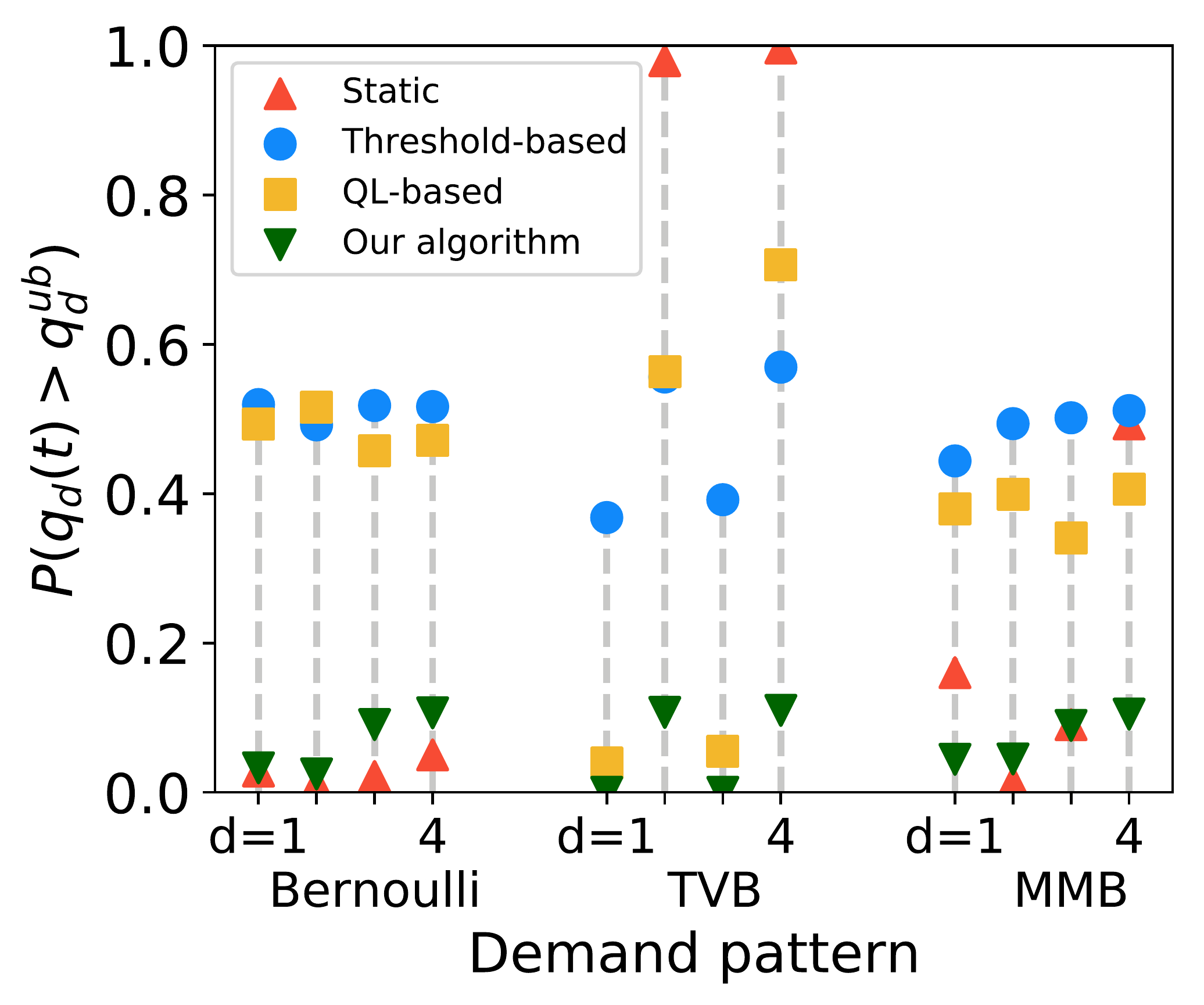}
    \caption{Violation probability $\text{P}(q_d~\geq~q^{ub}_d)$ for each PDC.}
    \label{fig:Pr_all}
\end{figure}

Although it seems that the static algorithm would be a more reasonable option in this case given its simplicity and better performance in most of the metrics, the advantage of the dynamic algorithms becomes apparent as we consider bursty arrivals with high and low demand periods. Since this is very unlikely to have simultaneous peak demands at all the PDCs, the dynamic algorithms can satisfy the QoS constraints with less number of UAVs in contrast to the static assignment, which leads to inefficient use of resources. This concept is often referred to as \emph{statistical multiplexing}.

\subsubsection{Time-Varying Bernoulli Arrivals (TVB)} In many real-world applications, the arrival rates can change drastically over time and therefore can create high demand and low demand periods. 
While a static resource assignment can either lead to underuse or overuse of the resources in some PDCs, the dynamic algorithms can relocate the resources to match the demand in each PDC. This can be seen in Fig.~\ref{fig:Pr_all}, where the violation probabilities for the static algorithm are almost equal to $1$ in the 2nd and the 4th PDCs (saturated situation), while the other two PDCs have zero violation probabilities (low demand). On the other hand, the dynamic algorithms can stabilize all the PDCs. 

As shown in Table~III, our algorithm has the best performance compared to the other baselines in all the metrics. Moreover, it can guarantee a maximum violation probability of $10\%$, as desired. The variations of the queue length and the number of UAVs used by our controller are shown in Fig.~\ref{fig:n_q_2} for an arbitrary sample path of arrivals.
\subsubsection{Markov-Modulated Bernoulli Arrivals (MMB)} In contrast to the previous case where the transitions were deterministic and periodic, the transition between the high demand and the low demand intervals is governed by a Markov process in this experiment. While this is possible that all the PDCs have high (low) demands at the same time, the probability of this event is small compared to a balanced situation where some have high, and the rest have low demands. Again, we can observe from Table~I that our algorithm has the best performance among all the baselines such that $p^{\max}$ is bounded by $0.1$, while the average number of UAVs in use are around $54$. Fig.~\ref{fig:n_q_3} shows the adjustments of the number of UAVs by our algorithm for a given sample path of the arrivals.

\section{CONCLUSIONS}

We have developed an RL-based resource management algorithm for a drone delivery system. In our approach, the number of UAVs allocated to each PDC is dynamically adjusted to match the demand in the system. We have shown that our approach can provide probabilistic upper bounds on the queue length of the packages waiting in the PDCs. We have evaluated the performance of our proposed algorithm for three different package arrival distributions including Bernoulli (Poisson), Time-varying Bernoulli, and Markov-modulated Bernoulli processes. We have shown that our algorithm outperforms the baselines in terms of the number of resources it uses to guarantee the QoS requirement. In particular, for the complicated arrival patterns (Time-varying Bernoulli and Markov-modulated Bernoulli) where the baselines are unable to satisfy the QoS constraint, our algorithm guarantees the QoS requirement for all PDCs, while minimizing the average number of UAVs in use. 


\bibliographystyle{IEEEtran}
\bibliography{bibfile}

\end{document}